\documentclass[review]{elsarticle}

\usepackage{amsmath}
\usepackage{amssymb}
\usepackage{url}
\usepackage{caption}
\usepackage{subcaption}
\usepackage{wrapfig}
\usepackage{textcomp,gensymb}
\usepackage{color}
\usepackage{multirow}
\usepackage{enumitem}

\usepackage[export]{adjustbox}
\usepackage[ruled]{algorithm2e}
\usepackage{pythontex}
\newcommand{\model}{ADP\xspace}
\newcommand{\zeromodel}{ZS-ADP\xspace}
\newcommand{\ssmodel}{SS-ADP\xspace}

\journal{ArXiv}

\begin{document}

\begin{frontmatter}

\title{Generative Adversarial Data Programming}

\author[mymainaddress]{Arghya Pal\corref{mycorrespondingauthor}}
\fntext[mymainaddress]{Postal Address: Department of Computer Science \& Engineering, Indian Institute of Technology Hyderabad, Hyderabad, India}

\author[mymainaddress]{Vineeth N Balasubramanian}

\cortext[mycorrespondingauthor]{Corresponding author: Arghya Pal, email address: arghyapal5@gmail.com}


\begin{abstract}
The paucity of large curated hand-labeled training data forms a major bottleneck in the deployment of machine learning models in computer vision and other fields. Recent work (Data Programming) has shown how distant supervision signals in the form of labeling functions can be used to obtain labels for given data in near-constant time. In this work, we present Adversarial Data Programming (ADP), which presents an adversarial methodology to generate data as well as a curated aggregated label, given a set of weak labeling functions. More interestingly, such labeling functions are often easily generalizable, thus allowing our framework to be extended to different setups, including self-supervised labeled image generation, zero-shot text to labeled image generation, transfer learning, and multi-task learning.
\end{abstract}


\end{frontmatter}

\section{Introduction}
\label{sec_introduction}
Curated labeled data is a key building block of modern machine learning algorithms and a driving force for deep neural network models that work in practice. However, the creation of large-scale hand-annotated datasets in every domain is a challenging task due to the requirement for extensive domain expertise, long hours of human labor and time - which collectively make the overall process expensive and time-consuming. Even when data annotation is carried out using crowdsourcing (e.g. Amazon Mechanical Turk), additional effort is required to measure the correctness (or goodness) of the obtained labels. We seek to address this problem in this work. In particular, we focus on automatically learning the parameters of a given joint image-label probability distribution (as provided in training image-label pairs) with a view to automatically create the labeled dataset.

\begin{figure}[!htb]
    \centering
    \includegraphics[width=\textwidth,height=0.5\textwidth]{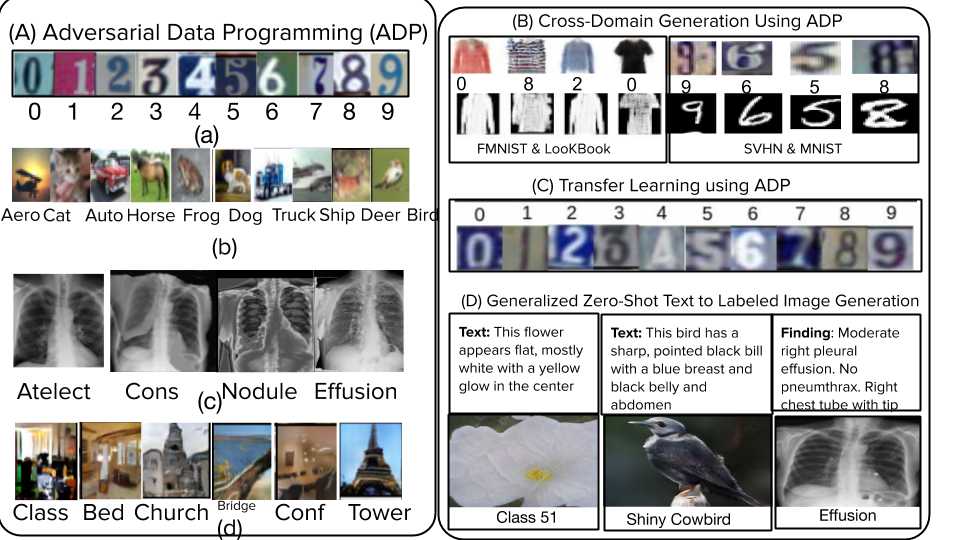}
    \caption{\textbf{(A) Labeled image generations of \model on different datasets:} (a)-(d) Sample results of image-label pairs generated using the proposed \model framework on SVHN \cite{SVHN}, CIFAR 10 \cite{CIFAR10}, Chest-Xray-14 \cite{CHEST}, and LSUN \cite{LSUN} datasets (respectively);  \textbf{(B) Cross-domain labeled image generation using \model:}; \textbf{(C) Transfer learning using \model:} \model transfers its knowledge from a source domain to a target domain if distant supervision signals are common. We demonstrate transfer learning from MNIST \cite{MNIST} to SVHN; \textbf{(D) Generalized zero-shot text to labeled image generation:} \zeromodel is a first-of-its-kind model that performs zero-shot text to labeled image generation on Flower102 \cite{Nilsback08}, UCSD Bird \cite{WahCUB_200_2011} and Chest-Xray-14 \cite{CHEST} datasets.}
    \label{fig:Motivational_example}
\end{figure}

\begin{figure}[!htb]
\centering
    \includegraphics[width=\textwidth, height=0.55\textwidth]{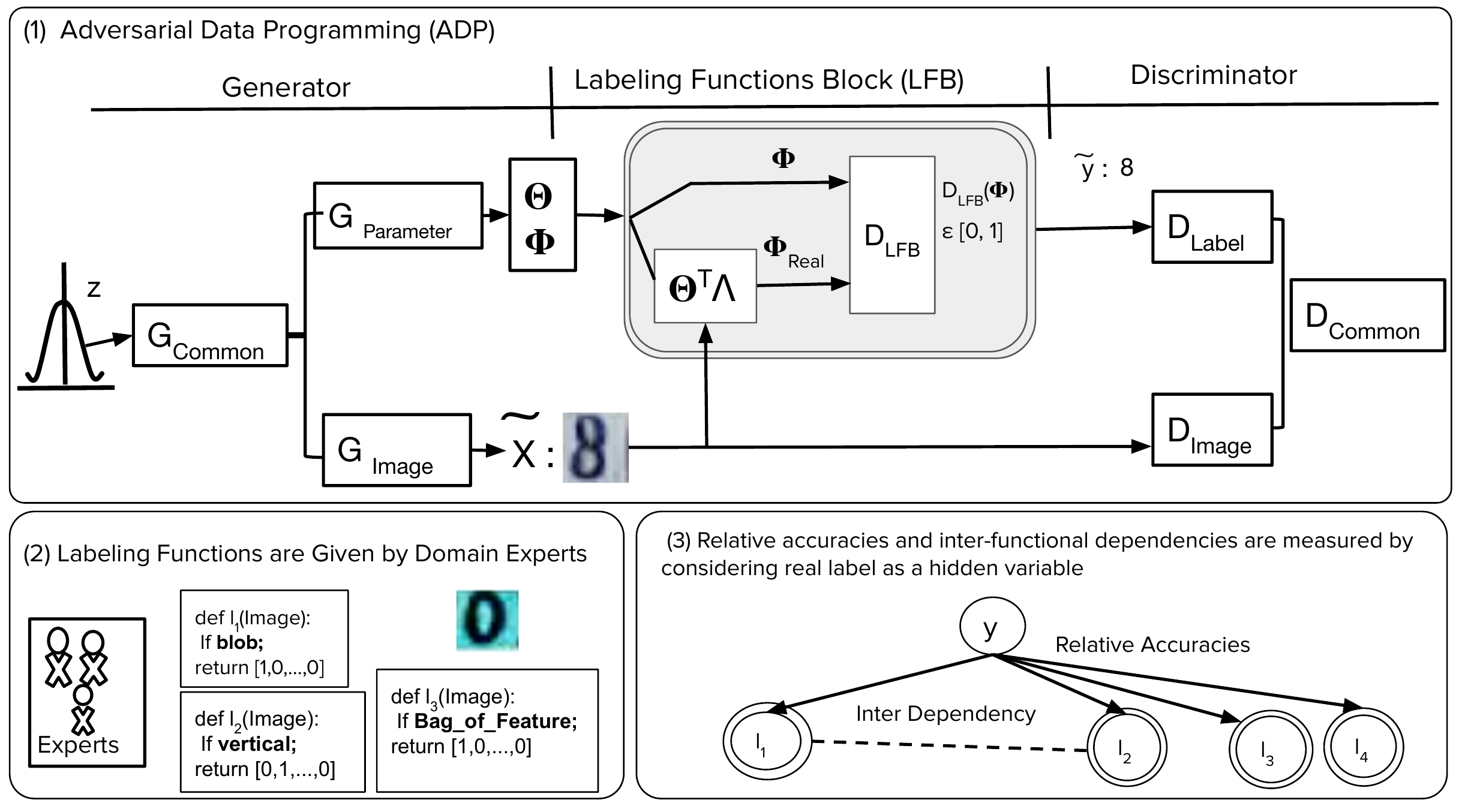}
    \caption{\textbf{(1)} Overall architecture of the \textit{Adversarial Data Programming (ADP) framework}. The generator generates image $X$ and parameters ($\Theta, \Phi$) which are used by the labeling functions block (LFB) to generate labeled images ($X, y$); \textbf{(2)} \textit{Labeling Functions:} Crowd experts give distant supervision signals in the form of weak labeling functions: e.g. presence of blob; \textbf{(3)} \textit{Dependencies of Labeling Functions:} Labeling functions show ``relative accuracies'' (solid arrow) and ``inter-functional dependencies'' (dotted line). The \model encapsulates all labeling functions in a unified abstract container called LFB. LFB helps learn parameters corresponding to both kinds of dependencies: ``relative accuracies'' and ``inter-functional dependency'' to generates labeled images.}
    \label{fig:Diagram}
\end{figure}

To this end, we exploit the use of distant supervision signals to generate labeled data. These distant supervision signals are provided to our framework as a set of weak labeling functions which represent domain knowledge or heuristics obtained from experts or crowd annotators. This approach has a few advantages: (i) labeling functions (which can even be just loosely defined) are cheaper to obtain than collecting labels for a large dataset; (ii) labeling functions act as an implicit regularizer in the label space, thus allowing good generalization; (iii) with a small fine-tuning, labeling functions can be easily re-purposed for new domains (\textit{transfer learning}) and multi-task learning (discussed in Section \ref{subsec:transfer_learning}); and (v) the labeling functions can be generalized to using semantic attributes, which aids adapting the approach of generalized zero-shot text to labeled image generation \ref{subsec:zero_shot}. We note that, writing a set of labeling functions (as we found in our experiments) is fairly easy and quick - we demonstrate three python-like functions as labeling functions to weakly label the SVHN \cite{SVHN} digit ``0'' in Figure \ref{fig:Diagram}.2.a (def l$_{1}$ - def l$_{3}$). Figure \ref{fig:Motivational_example} shows a few examples of our results to illustrate the overall idea.

In practice, labeling functions can be associated with two kinds of dependencies: (i) relative accuracies (shown as solid arrows in Figure \ref{fig:Diagram}.2.b), are weights to labeling functions measuring the correctness of the labeling functions w.r.t. the true class label ; and (ii) inter-function dependencies (shown as dotted line in Figure \ref{fig:Diagram}.2.b) that capture the relationships between the labeling functions concerning the predicted class label. In this work, we propose a novel adversarial framework, i.e. Adversarial Data Programming (\model) presented in Figure \ref{fig:Diagram}.1, using Generative Adversarial Network (GAN) that learns these dependencies along with the data distribution using a minmax game. 

Our broad idea of learning relative accuracies and inter-function dependencies of labeling functions is inspired by the recently proposed Data Programming (DP) framework \cite{ratner2016data} (and hence, the name ADP), but our method is different in many ways: (i) DP learns $P(\tilde{y}|X)$), while \model learns joint distribution, i.e. $P(X,y)$; (ii) DP uses Maximum Likelihood Estimation (MLE) to estimate relative accuracies of labeling functions. We instead use adversarial framework to estimate relative accuracies and inter-function dependencies of labeling functions. We note that, \cite{danihelka2017comparison} and \cite{theis2015note} provide insights on the advantage of using a GAN-based estimator over MLE. (iii) We use adversarial approach to learn inter functional dependencies of labeling functions and replaces the computationally expensive factor graph modeling proposed in \cite{ratner2016data}.

Furthermore, we show applicability of \model in different tasks, such as: (i) Self supervised labeled image generation (\ssmodel), that generates labeled image using an unlabeled dataset. The \ssmodel dependencies of labeling functions using image rotation based self-supervised loss (similar to the image rotation loss proposed in \cite{chen2019self}). (ii) Generalized zero-shot text to labeled image synthesis (\zeromodel) that generates labeled images from textual descriptions (see Section \ref{subsec:zero_shot}). We show that \zeromodel infer zero-shot classes as well as seen classes of generated images using labeling functions those are semantic attributes (similar to semantic attributes proposed by Lampert \textit{et al.} \cite{lampert2009learning}). To the best of our knowledge, the \zeromodel is the first generalized zero-shot text to labeled image generator.

As outcomes of this work, we show a framework to integrate labeling functions within a generative adversarial framework to model joint image label distribution. To summarize:
\begin{itemize}
\item We propose a novel adversarial framework, \model, to generate robust data-label pairs that can be used as datasets in domains that have little data and thus save human labor and time.
\item The proposed framework can also be extended to incorporate generalized zero-shot text to labeled image generation, i.e. \zeromodel; and self-supervised labeled image generation, i.e. \ssmodel in Section \ref{sec_extendability}. 
\item We demonstrate that the proposed framework can also be used in a transfer learning setting, and multi-task joint distribution learning where images from two domains are generated simultaneously by the model along with the labels, in Section \ref{sec_extendability}.
\end{itemize}
\vspace{-0.3cm}
\section{Related Work}
\label{sec:related_work}
\vspace{-0.3cm}
\noindent \textbf{Distant Supervision:} In this work, we explored the use of distant supervision signals (in the form of labeling functions) to \textit{generate} labeled data points. Distant supervision signals such as labeling functions are cheaper than manual annotation of each data point, and have been successfully used in recent methods such as \cite{ratner2016data, fries2019weakly}. MeTaL \cite{ratner2018snorkel} extends \cite{ratner2017learning} by identifying multiple sub-tasks that follow a hierarchy and then provides labeling functions for sub-tasks. These methods require unlabeled test data to generate a labeled dataset and computationally expensive due to the use of Gibbs sampling methods in the estimation step (also shown in our results).\\
\noindent \textbf{Learning Joint Distribution using Adversarial Methods:}
In this work, we use an adversarial approach to learn the joint distribution by weighting a set of domain-specific label functions using a Generative Adversarial Network (GAN). We note efforts \cite{pu2018jointgan, lucic2019high} which attempt to train GANs to sample from a joint distribution. In this work, we propose a novel idea to instead use distant supervision signals to accomplish learning the joint distribution of labeled images, and compare against these methods.\\
\noindent \textbf{Generalized Zero-shot Learning:}
A typical \textit{generalized zero-shot model} (such as \cite{zhang2018model,liu2018generalized,zhang2019co}) learns to transfer learned knowledge from seen to unseen classes by learning correlations between the classes at training time, and recognizing both seen and unseen classes at test time. While, efforts such as \cite{reed2016generative, zhang2016stackgan, zhang2017stackgan++, xu2018attngan, zhang2018photographic} proposed text-to-image generation methodology and then demonstrated results on zero-shot text to image generation. However, no work has studied the problem of \textit{generalized zero-shot text-to-labeled-image} generation so far. We integrate a set of semantic attribute-based distant supervision, similar to proposed by Lampert \textit{et al.} \cite{lampert2009learning}, signals such as color, shape, part etc. to identify seen and zero-shot visual class categories.\\
\noindent \textbf{Self Supervised Labeled Image Generation}: While self supervised learning is an active area of research, we found only one work \cite{lucic2019high} that performs self supervised labeled image generation. In particular, \cite{lucic2019high} uses a GAN framework that performs k-means cluster within discriminator and does an unsupervised image generation. In this work, we instead use a set of labeling functions and perform self supervision. We defer the discussion to Section \ref{sec_extendability}.
\section{Adversarial Data Programming: Methodology}
\label{sec_adp_methodology}
\noindent Our central aim in this work is to learn parameters of a probabilistic model:
\begin{equation}
\label{eq:joint_distribution}
P(X,y|z)    
\end{equation}
that captures the joint distribution over the image $X$ and the corresponding labels $y$ conditioned on a latent variable $z$.

To this end, we encode distant supervision signals as a set of (weak) definitions by annotators using which unlabeled image can be labeled. Such distant supervision signals allow us to weakly supervise images where collection of direct labeled image is expensive, time consuming or static. We encapsulate all available distant supervision signals, henceforth called \textit{labeling functions}, in a unified abstract container called \textbf{L}abeling \textbf{F}unctions \textbf{B}lock (LFB, see Figure \ref{fig:Diagram}.1). Let LFB comprised of $n$ labeling functions $l_1, l_2,\cdots, l_n$, where each labeling function is a mapping, i.e.: $l_i: X_j  \rightarrow \textbf{a}_{ij}$, that maps an unlabelled image $X$ to a class label vector, $\textbf{a}_{ij} \in \mathbb{R}^m$, where $m$ is the number of classes labels with $\sum_{k=1}^{m} a_{ij}^{k} = 1 \text{ and, } 0 \leq a_{ij}^k \leq 1$. For example, as shown in Figure \ref{fig:Diagram}.2, $X_j$ could be thought of as an image from the CIFAR 10 single digit dataset, and $\textbf{a}_{ij} \in \mathbb{R}^{10}$ would be the corresponding label vector when the labeling function $l_i$ is applied on $X_j$. The $\textbf{a}_{ij}$, for instance, could be the one-hot 10-dimensional class vector, see Figure \ref{fig:Diagram}.2.

We characterize the set of labeling functions, $ L = \{l_1, l_2, \cdots, l_n\}$ as having two kinds of dependencies: (i) \textit{relative accuracies} are weights given to labeling functions based on whether their outputs \textit{agree} with true class label $y$ of an image $X$; and (ii) \textit{inter-function dependencies} capture the relationships between the labeling functions with respect to the predicted class label. To obtain a final label $y$ for a given data point $X$ using the LFB, we use two different sets of parameters, $\Theta$ and $\Phi$ to capture each of these dependencies between the labeling functions. We, hence, denote the Labeling Function Block (LFB) as:
\begin{equation} 
\label{eq:labeling_functions_block}
LFB_{L,\Theta,\Phi}: X_j \rightarrow \textbf{a}_{j}
\end{equation}
i.e. given a set of labeling functions $L$, a set of parameters capturing the relative accuracy-based dependencies between the labeling functions, $\Theta$, and a second set of parameters capturing inter-label dependencies, $\Phi$, $LFB$ provides a probabilistic label vector, $\textbf{a}_j$, for a given data input $X_j$.

Our Equation \ref{eq:joint_distribution} that we seek to model in this work (Equation \ref{eq:joint_distribution}) hence becomes:
\begin{equation} 
\label{eq:modified_joint_distribution}
P(X, LFB_{L,\Theta,\Phi}(X)|z)
\end{equation}
In the rest of this section, we show how we can learn the parameters of the above distribution modeling image-label pairs using an adversarial framework with a high degree of label fidelity. We use Generative Adversarial Networks (GANs) to model the joint distribution in Equation \ref{eq:modified_joint_distribution}. In particular, we provide a mechanism to integrate the LFB (Equation \ref{eq:labeling_functions_block}) into the GAN framework, and show how $\Theta$ and $\Phi$ can be learned through the framework itself. Our adversarial loss function is given by:
\begin{equation} 
\label{eq:base_adversarial_function}
\begin{split}
& \min \max L(G,D) = \\
& \mathbb{E}_{(X,y) \sim \mathcal{D}} \log(D(X, y)) + \mathbb{E}_{(\tilde{X}, \Theta, \Phi) \sim P_{fake}(z)} \log(1-D(\tilde{X}, LFB_{L, \Theta, \Phi}(\tilde{X}))
\end{split}
\end{equation}
where $G$ is the generator module and $D$ is the discriminator module. The overall architecture of the proposed ADP framework is shown in Figure \ref{fig:Diagram}.1.
\subsection{The Proposed Framework}
\label{sec_framework}
\noindent A. \textbf{\model Generator} $G(\widetilde{X},\Theta,\Phi|z)$: Given a noise input $\textbf{z}\sim\mathcal{N}(0,I)$ and a set of $n$-labeling functions $L$, the generator $G(.)$ outputs an image $X$ and the parameters $\Theta$ and $\Phi$, the dependencies between the labeling functions described earlier. In particular, $G(.)$ consists of three blocks: $G_{Common}$, $G_{Image}$ and $G_{Parameter}$, as shown in Figure \ref{fig:Diagram}.1. $G_{Common}$ captures the common high-level semantic relationships between the data and the label space, and is comprised only of fully connected (FC) layers. The output of $G_{Common}$ forks into two branches: $G_{Image}$ and $G_{Parameter}$, where $G_{Image}$ generates the image $\tilde{X}$, and $G_{Parameter}$ generates the parameters $\{\Theta, \Phi\}$. While $G_{Parameter}$ uses FC layers, $G_{Image}$ uses Fully Convolutional (FCONV) layers to generate the image (more details in Section \ref{sec_expts}). Figure \ref{fig:Diagram}.1 also includes a block diagram for better understanding.

\noindent B. \textbf{\model Discriminator} $D(X, y)$: The discriminator $D(.)$ estimates the likelihood of an image-label input pair being drawn from the real distribution obtained from training data. The $D(.)$ takes a batch of either real or generated image and inferred label (from $LFB$) pairs as input and maps that to a probability score to estimate the aforementioned likelihood of the image-label pair. To accomplish this, $D(.)$ has two branches: $D_{Image}$ and $D_{Label}$ (shown in the Discriminator part in Figure \ref{fig:Diagram}.1). These two branches are not coupled in the initial layers, but the branches share weights in later layers and become $D_{Common}$ to extract joint semantic features that help $D(.)$ classify correctly if an image-label pair is \textit{fake} or \textit{real}. 
\begin{algorithm}[ht]
\SetAlgoLined
\textbf{Input:} Labeling functions $\{l_1,\cdots,l_n\}$, Relative accuracies $\theta_1,\cdots,\theta_n$, Output probability vectors of labeling functions $\textbf{a}_1,\cdots,\textbf{a}_n$ of $n$-generated images by $G$\\
\textbf{Output:} $\Phi_{Real}$

Set $\Phi_{Real} = \text{I}(n,n)$\; \tcc{\footnotesize{I = Identity Matrix}}
 
\For{$i = 1$ to $n$}{ \tcc{\footnotesize{For each labeling function}}

  \For{$j = i+1$ to $n$}{ \tcc{\footnotesize{For each other labeling function}}
  
     \tcc{\footnotesize{If one-hot encoding of the outputs of two functions match, increment $(i,j)$th entry in $\Phi_{Real}$ by 1}}
     $\Phi_{Real}(i,j) = \Phi_{Real}(i,j) + \text{OneHot}(\theta_i \textbf{a}_i) \cdot \text{OneHot}(\theta_j \textbf{a}_j)$\;
  }
 }

\For{$p = 1$ to $n$}{
 $\Phi_{Real}$(p, .) = $\frac{\Phi_{Real}(p, .)}{\sum_{u=1}^n \Phi_{Real}(p,u)}$\;
}

Set $\Phi_{Real} = \Phi_{Real} + \Phi_{Real}^T - \text{diag}(\Phi_{Real})$ \tcc{\footnotesize{Complete using symmetry}}
\caption{Procedure to compute $\Phi_{Real}$}
\label{alg:computing_real_phi}
\end{algorithm}

\noindent C. \textbf{Labeling Functions Block} $LFB(\tilde{y}|\tilde{X}, \Theta, \Phi)$: This is a critical module of the proposed ADP framework. Our initial work revealed that a simple weighted (linear or non-linear) sum of the labeling functions does not perform well in generating out-of-sample image-label pairs. We hence used a separate adversarial methodology within this block to learn the dependencies. We describe the components of the LFB below.\\
\textit{C.1. Relative Accuracies, $\Theta$, of Labeling Functions}: In this, we assume that all the labeling functions infer label of an image independently (i.e. independent decision assumption) and the parameter $\Theta$ gives relative weight to each of the labeling functions based on their correctness of inferred label for true class $y$. The output, $\Theta$, of the $G_{Parameter}$ block in the ADP Generator $G(.)$, provides the relative accuracies of the labeling functions. Given the $j^{th}$ image output generated by $G_{Image}$: $\tilde{X}_{j}$, the $n$-labeling functions $\{l_1, l_2, \cdots, l_n\}$, and the probabilistic label vectors $\{\textbf{a}_{ij},i=1,\cdots,n\}$ obtained using the labeling functions, we define the aggregated final label as:
\begin{equation} 
\label{eq:learning_theta}
\tilde{y}_{j} = \sum_{i=1}^{n}\tilde{\theta}_{i} \textbf{a}_{ij}\\
= \tilde{\Theta}\cdot \textbf{a}_{j}
\end{equation}
\begin{algorithm}
\textbf{Input:} Iterations: $N$, Number of steps to train $D$: $k$, $m$\\
\textbf{Output:} Trained ADP model\\
\For{N}{
    \For{$k$ steps}{
        Draw $m$ samples from G: $\{(\tilde{X}_1, \Theta_1, \Phi_1),\cdots, (\tilde{X}_m, \Theta_m, \Phi_m)\}$ and subsequently infer corresponding labels $\{\widetilde{\textbf{y}}_1,\cdots,\widetilde{\textbf{y}}_m\}$ using LFB(.)(Equation \ref{eq:learning_phi})\;
        Update weights of $D$ and $D_{LFB}$ ($\psi_{d}$ and $\psi_{l}$ respectively):
        \[\nabla_{\psi_{d}} \frac{1}{m}\sum_{i=1}^{m}\Big[\log D(X_i, y_i) + \log (1- D(\tilde{X}_i, \tilde{y}_i))\Big]\]
        \[\nabla_{\psi_{l}} \frac{1}{m}\sum_{i=1}^{m}\Big[\log D_{LFB}(\Phi_{real_i}) + \log (1 - D_{LFB}(\Phi_i))\Big]\]
        }
    Update weights of generator $G$ (i.e. $\psi_{g}$):\;
        \[\nabla_{\psi_{g}} \frac{1}{m}\sum_{i=1}^{m}\Big[\log (1 - D_{LFB}(\Phi_i))\Big]\]
        \[\nabla_{\psi_{g}} \frac{1}{m}\sum_{i=1}^{m}\Big[\log (1- D(\tilde{X}_i, \tilde{y}_i))\Big]\]
}
\caption{Training procedure of \model}
\label{alg:main}
\end{algorithm}
where $\tilde{\theta}_{i}$ is the normalized version of $\theta_{i}$, i.e. $\tilde{\theta}_{i} = \frac{\theta_{i}}{\sum_{k=1}^n \theta_{k}}$. The aggregated label, $\tilde{y}$, is provided as an output of the LFB.\\
\textit{C.2. Inter-function Dependencies, $\Phi$, of labeling functions}: In practice, a dependency among labeling function is a common observation. Studies \cite{ratner2016data, fries2019weakly} show that such dependencies among labeling function proportionally increase with the number of labeling functions. Modeling such inter-functional dependencies act as an implicit regularizer in the label space leading to an improvement in the labeled image generation quality and generated image-to-label correspondence. 

While recent studies utilized factor graph \cite{ratner2016data,fries2019weakly} to learn such dependencies among labeling functions, we instead use an adversarial mechanism inside the LFB to capture inter-function dependency $\tilde{\Phi}$ that in turns influence the final relative accuracies, $\tilde{\theta}$. $D_{LFB}$, a discriminator inside LFB, receives two inputs: $\Phi$, which is output by $G_{Parameter}$, and $\Phi_{Real}$, which is obtained from $\Theta$ using the procedure described in Algorithm \ref{alg:computing_real_phi}. Algorithm \ref{alg:computing_real_phi} computes a matrix of interdependencies between the labeling functions, $\Phi_{Real}$, by looking at the one-hot encodings of their predicted label vectors. If the one-hot encodings match for given data input, we increase the count of their correlation. The task of the discriminator is to recognize the computed interdependencies as real, and the $\Phi$ generated through the network in $G_{Parameter}$ as fake. The objective function of our second adversarial module is hence:
\begin{equation} 
\label{eq:second_adversarial_function}
\begin{split}
\min \max L(G, D_{LFB}) =  \mathbb{E}\log(D_{LFB}(\Phi_{real}(\textbf{z}))) + \mathbb{E}\log(1-D_{LFB}(\Phi(\textbf{z})))
\end{split}
\end{equation}
\noindent where $\Phi_{Real}$ and $\Phi$ are obtained from $G_{Parameter}(z)$ as described above. More details of the LFB are provided in implementation details in Section \ref{sec_expts}.\\
\textit{C.3. Final label prediction, $\tilde{y}$, using LFB}: We define the aggregated final label as:
\begin{equation} 
\label{eq:learning_phi}
\tilde{y}_{j} = \tilde{\Theta}_{j}\cdot \Phi^{T}_{j}\cdot \textbf{a}_{j}
\end{equation}
The samples $(\tilde{X},\tilde{y})$ generated using the $G$ and $LFB$ modules thus provide samples from the desired joint distribution (Eqn \ref{eq:joint_distribution}) modeled using the framework.
\subsection{Final Objective Function}
\label{sec_objective_function}
We hence expand our objective function from Equation \ref{eq:base_adversarial_function} to the following:
\begin{equation} 
\label{eq:modified_adversarial_function}
\begin{split}
& \min \max L(G,D_{image}, D_{label}) = \\ & \mathbb{E}_{(X,y) \sim \mathcal{D}} \log(D_{image}(X)) + \mathbb{E}_{\textbf{z} \sim \mathcal{N}(0,I)}\log(1 - D_{image}(G_{image}(z)))\\
& + \mathbb{E}_{(X,y) \sim \mathcal{D}} \log(D_{label}(y)) + \mathbb{E}_{\textbf{z} \sim \mathcal{N}(0,I)}\log(1 - (D_{label}(LFB(G(z))))
\end{split}
\end{equation}
\subsection{Training}
\label{sec_training}
Algorithm \ref{alg:main} presents the overall stepwise routine of the proposed \model method. During the training phase, the algorithm updates weights of the model by estimating gradients for a batch of labeled data points. 
\section{Theoretical Analysis}
\label{sec_theoretical_analysis}
\noindent\textbf{Theorem:}\textit{
For any fixed generator $G$, the optimal discriminator $D$ of the game defined by the objective function $L(G, D)$ is
\begin{equation} 
\label{eq:third_adversarial_function}
\begin{split}
D^*(X, y) = \frac{p_{real} (X, y)}{p_{real}(X,{y}) + p_{fake}(X, y)}
\end{split}
\end{equation}
}
\noindent\textbf{Proof:}
The training criterion for the discriminator D, given any generator $G$, is to maximize the quantity $L(G, D)$. Following \cite{goodfellow2016nips}, maximizing objective function $L(G, D)$ depends on the result of Radon-Nikodym Theorem \cite{donahue2016adversarial}, i.e.
\begin{equation} 
\label{eq:Radon-Nikodym_Theorem}
\begin{split}
\mathbb{E}_{\textbf{z} \sim p_{fake}(\textbf{z})}\log(1-D(G(\textbf{z}))) = \mathbb{E}_{x \sim p_{real}(x)}\log(1-D(x))
\end{split}
\end{equation}
\noindent
The objective function can be reformulated for ADP as:
\begin{equation}
\label{eq:ADP_Radon-Nikodym_Theorem}
\begin{split}
L(G, D) & \equiv \quad \int_{y} L_y(G,D) dy
\end{split}
\end{equation}

Following Radon-Nikodym Theorem we can say:
\begin{equation}
\begin{split}
& V_y(G,D) = \int_{X} p_{data}(X, y) \log(D(X, y)) dX + \int_{\textbf{z}} p_{fake}(\textbf{z}) \log(1 - D(G(\textbf{z})) d\textbf{z}\\
& + \int_{X}  p_{fake}(X, y) \log(1-D(X, y)) dX\\
& = \quad \int_{X} \bigg[ p_{data}(X, y) \log(D(X, y)) + p_{fake}(X, y) \log(1-D(X, y)) \bigg] dX
\end{split}
\end{equation}
Now, from Equation \ref{eq:ADP_Radon-Nikodym_Theorem}:
\begin{equation}
\begin{split}
L(G,D) & \equiv \quad \int_{y} L_y(G,D) dy\\
& = \quad \int_{X} \int_{y} \bigg[ p_{data}(X, y) \log(D(X, y)) + \\
& p_{fake}(X, y) \log(1-D(X, y)) \bigg] dydX
\end{split}
\end{equation}

Following \cite{goodfellow2016nips}, for any $(a,b) \in \mathbb{R}^2 \backslash \{0, 0\}$, the function $y \rightarrow a\log y + b\log(1 - y)$ achieves its maximum in $[0,1]$ at $\frac{a}{a+b}$, which proves the claim.

\noindent\textbf{Theorem:}\textit{
The equilibrium of V(G, D) is achieved if and only if $p_{data}(X, y) = p_{g}(X, y)$, and $\max_{D} L(G, D)$ attains the value $-\log(4)$.
}

\noindent\textbf{Proof:} Considering the optimal discriminator $D^{*}(X, y)$ and the fixed generator $G$ described in Eqn \ref{eq:second_adversarial_function} in Theorem 1, the min-max game in Eqn \ref{eq:base_adversarial_function} can be reformulated as:
\begin{equation}
\begin{split}
& C(G) = \max_{D} L(G, D) = \int_{x} \int_{y} \bigg[ p_{data}(X, y) \log(D(X, y))\\ 
& + p_{fake}(X, y) \log(1-D(X, y)) \bigg] dydX\\
& = \mathbb{E}_{(X, y) \sim p_{real}} \big[\log D_{G}^{*}(X, y)] + \mathbb{E}_{(X, y) \sim p_{fake}} \big[\log D_{G}^{*}(X, y)]\\
& = \mathbb{E}_{(X, y) \sim p_{real}} \Bigg[\log \frac{p_{real} (X, y)}{p_{real}(X,{y}) + p_{fake}(X, y)}\Bigg]\\
& + \mathbb{E}_{(X, y)\sim p_{fake}} \Bigg{[}\log \frac{p_{fake} (X, y)}{p_{real}(X,{y}) + p_{fake}(X, y)}\Bigg{]}
\end{split}
\end{equation}

The training criterion reaches its global minimum $p_{real}(X, y) = p_{fake}(X, y)$ and at this point, $D_{G}^{*}(X, y)$ will have the value $\frac{1}{2}$. We hence have:
\begin{equation}
\begin{split}
& C(G) = \mathbb{E}_{(X, y) \sim p_{real}} \Bigg[\log \frac{p_{real} (X, y)}{p_{real}(X,{y}) + p_{fake}(X, y)}\Bigg] \\
& + \mathbb{E}_{(X, y) \sim p_{fake}} \Bigg[\log \frac{p_{fake} (X,y)}{p_{real}(X,{y}) + p_{fake}(X, y)}\Bigg] \\
& = \mathbb{E}_{(X, y) \sim p_{real}} \bigg[ \log \frac{1}{2} \bigg] + \mathbb{E}_{(X, y) \sim p_{fake}} \bigg[ \log \frac{1}{2} \bigg] = - \log 4
\end{split}
\end{equation}
So, with a \textit{fixed} generator $G$, the training criterion $C(G)$ attain its best possible value when $p_{real}(X, y) = p_{fake}(X, y)$. At training phase, the criterion $C(G_1)$ uses generator $G_1$ and optimizes the objective function $L(G,D)$
\begin{align*}
C(G_1) 
& = - \log 4 + KL\Bigg( p_{real} \Bigg|\Bigg| \frac{p_{real} + p_{fake}}{2}\Bigg) \\
& + KL\Bigg( p_{fake} \Bigg|\Bigg| \frac{p_{real} + p_{fake}}{2}\Bigg) \geq -\log4\\
& = - \log 4 + 2\cdot JSD\bigg(p_{real}||p_{fake}\bigg) \geq -\log4
\end{align*}
in this equation, KL denotes the Kullback-Leibler divergence and JSD denotes the Jensen-Shannon divergence. From the property JSD, it results non-negative if $p_{real}(X, y) \neq p_{fake}(X, y)$ and zero when $p_{real}(X, y) = p_{fake}(X, y)$. So, the global minimum is attained by the training criterion $C(G)$ at the $p_{real}(X, y) = p_{fake}(X, y)$. At that point, the criterion $C(G)$ attains the value $C(G) = - \log 4$ is the global minimum of $C(G)$ and at that point the generator perfectly mimics the real joint data-label distribution.
\section{Extensibility of ADP in different tasks}
\label{sec_extendability}
\subsection{Transfer Learning using \model}
\label{subsec:transfer_learning}
\begin{figure}
\centering
\includegraphics[width=0.7\textwidth, height=0.35\textwidth]{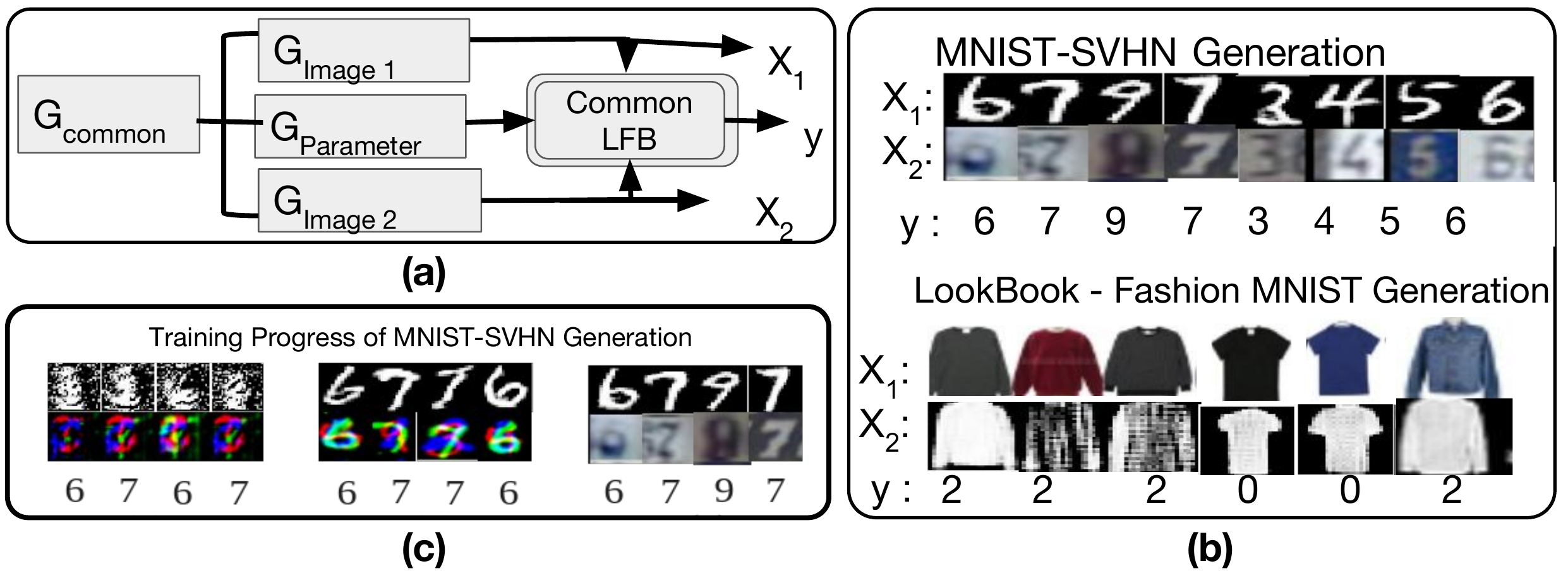}
\caption{\textbf{Multi-task joint distribution learning: (a) A modified \model can generates images of two domains $X_{1}$} and $X_{2}$ having a label correspondence, (b) labeled images from MNIST-SVHN and LookBook-FMNIST, (c) Training progress of MNIST-SVHN joint labeled image generation.}
\label{fig:multi_task_joint_distribution}
\end{figure}

Distant supervision signals such as labeling functions (which can often be generic) allows us to extend the proposed \model to a transfer learning setting. In this setup, we trained \model initially on a source dataset and then finetuned the model to a target dataset, with very limited training. In particular, we first trained \model on the MNIST dataset, and subsequently finetuned the $G_{image}$ branch alone on the SVHN dataset. We note that the weights of $G_{common}$, $G_{parameter}$ and $D_{LFB}$ are unaltered. The final finetuned model is then used to generate image-label pairs (which we hypothesize will look similar to SVHN). Figure \ref{fig:Motivational_example}C (named ``Transfer Learning using \model'') shows encouraging results of our experiments in this regard.
\subsection{Multi-task Joint Distribution Learning}
\label{subsec:multi_task_learning}
Learning a cross-domain joint distribution from heterogeneous domains is a challenging task. We show that the proposed \model method can be used to achieve this, by modifying its architecture as shown in Figure \ref{fig:multi_task_joint_distribution}(a) (top), to simultaneously generate data from two different domains, i.e. $X_{1}$ and $X_{2}$. We study this architecture on the: (1) MNIST and SVHN datasets, as well as (2) LookBook and Fashion MNIST datasets; and show promising results of our experiments in Figure \ref{fig:multi_task_joint_distribution}(b). The LFB acts as a regularizer and maintains the correlations between the domains in this case. We show joint multi-task joint distribution training progress on MNIST and SVHN datasets in Figure \ref{fig:multi_task_joint_distribution}(c).
\subsection{Self Supervised Labeled Image Generation}
\label{sec_distant_supervised}
\begin{figure}
\centering
\includegraphics[width=0.5\textwidth, height=0.25\textwidth]{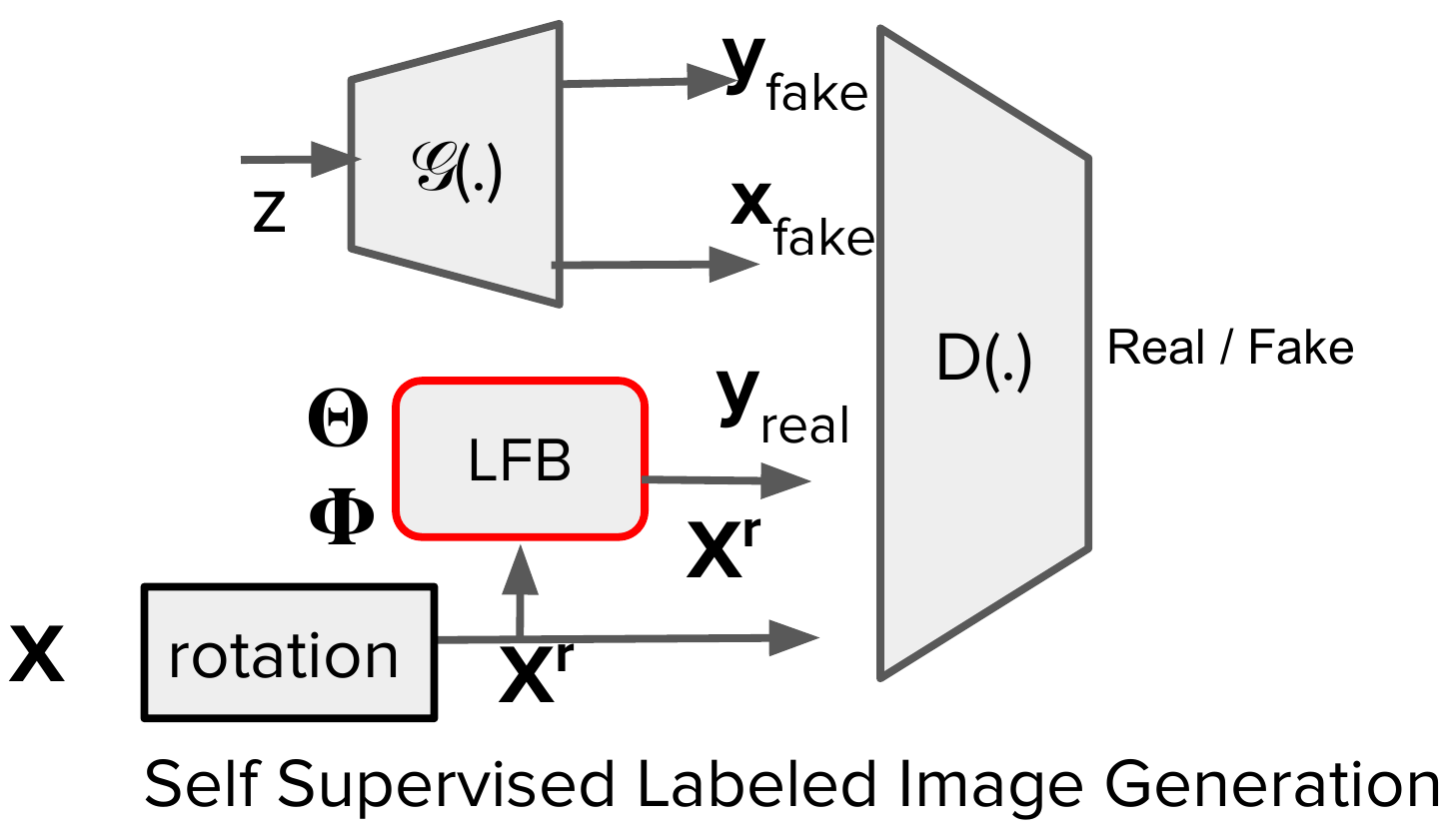}
\caption{\textbf{Block Diagram of \ssmodel}:}
\label{fig:self_supervised_LIG}
\end{figure}
Thus far, we show labeled image generation process using $LFB(.)$, which is integrated in the generator $G(.)$. In this section, we show the labeled image generation process from unlabeled data $\mathcal{D}_{u}$, and show a way to integrate $LFB(.)$ on real unlabeled image to get ``distant supervision'' image labels. In particular, $LFB(.)$ gets the real unlabeled image $X\sim\mathcal{D}_{u}$ as input and infers the label $y$. Similar to \cite{chen2019self}, we provide $LFB(.)$ a rotated version of image $X$, i.e. $LFB_{L,\Theta,\Phi}: X^{r}_j \rightarrow \textbf{a}_{j}$ and $y = \Theta\cdot \Phi^{T}\cdot \textbf{a}$, where $r\in\{0\degree, 90\degree, 180\degree, 270\degree\}$, see Figure \ref{fig:self_supervised_LIG}. The relative accuracy parameter $\Theta$ and inter-function dependency parameter $\Phi$ of $LFB(.)$ are learned based on how close the inferred label $y$ of unlabeled image $X$ to the inferred label $y^{r}$ of the rotation $X^{r}$ of image $X$, i.e:
\begin{equation}
    \min_{\Theta, \Phi} \mathcal{L}_{self} = \mathbb{E}_{\substack{y\sim LFB(X)\\y^{r}\sim LFB(X^{r})}}||y - y^{r}||_{1}
\end{equation}
Hence, the objective function of \ssmodel is as follows:
\begin{equation}
\begin{split}
& \min \max L(G,D_{image}, D_{label}) = \\
&\mathbb{E}_{X \sim \mathcal{D}_{u}} \log(D_{image}(X)) + \mathbb{E}_{\textbf{z} \sim \mathcal{N}(0,I)}\log(1 - D_{image}(G_{image}(z)))\\
& + \mathbb{E}_{X \sim \mathcal{D}_{u}} \log(D_{label}(LFB(X))) + \mathbb{E}_{\textbf{z} \sim \mathcal{N}(0,I)}\log(1 - (D_{label}(G(z)))
\end{split}
\label{eq:distant_supervised_adversarial_function}
\end{equation}
and the final objective is:
\begin{equation}
    \min \max L_{ss}(G,D_{image}, D_{label}) = L(G,D_{image}, D_{label}) + \lambda \mathcal{L}_{self}
\end{equation}
\subsection{Generalized Zero-shot Text to Labeled Image Synthesis using \model}
\label{subsec:zero_shot}
We go further to introduce a first-of-its-kind approach to generalized zero-shot text to labeled image generation, i.e. $p(X,LFB(X)|t)$, using a modified version of \model, henceforth called \zeromodel. Generalized zero-shot classification \cite{zhang2019co,liu2018generalized,zhang2018model} and text-to-(zero-shot)image synthesis $p(X|t)$ in \cite{reed2016generative,zhang2016stackgan,zhang2018photographic} were studied separately in literature, we go beyond text-to-image synthesis and propose a novel framework \zeromodel that performs \textit{text-to-labeled-image synthesis} in a generalized zero-shot setup. To accomplish such objective, we assume to have a dataset $\mathcal{D}_{G}$ with $K$ classes. Of $K$ classes, first $m$ seen classes have samples in the form of  tuples \{$X^{\text{seen}}, y^{\text{seen}}, t^{seen}\}$, where an image $X^{seen}$ is associated with a seen class $y^{seen}\in \{y_{1}, \cdots, y_{m}\}$, and $t^{seen}$ denotes textual description (such as caption) of image $X^{seen}$. Complementarily, we have no images for the rest of the classes, i.e. $y^{zero}\in\{y_{m+1}, \cdots, y_{K}\}$, and only textual descriptions $t^{zero}$ are available for zero-shot classes. Our primary aim is to learn the parameters of the following probabilistic model at training time:
\begin{equation}
    p(X^{seen}, LFB(X^{seen}) | t^{seen})
    \label{eq:zeroshot1}
\end{equation}
such that, at inference time, it can generate labeled images of both seen and zero-shot classes:
\begin{equation}
    p(X, LFB(X) | t)
    \label{eq:zeroshot2}
\end{equation}
\noindent where $t\in\{t^{seen} \cup t^{zero}\}$ and $X\in\{X^{seen} \cup X^{zero}\}$.

To this end, \zeromodel learns class-independent \textit{style}-ness information (background, illumination, object orientation). While, \zeromodel learns \textit{content}-ness information of visual appearances and attributes such as shape, color, and size etc. using $LFB$ module. We follow the semantic attribute based object class identification work of Lampert \textit{et al.} \cite{lampert2009learning}, and modify the labeling functions in $LFB$ in terms of a set of $p$ semantic attributes, i.e. $\textbf{s}= \{s_{1}, s_{2}, \cdots, s_{p}\}$. To make this exposition self-contained, we are paraphrasing the idea of ``identifying an class based on semantic attributes'' of \cite{lampert2009learning} using an example: a object ``zebra'' can be classified by recognizing semantic attributes, such as: ``four legs'', ``has tail'' and ``white-black stripes on body''. Such semantic attributes can be integrated within $LFB$ as labeling functions without any architectural change in \model (and hence \zeromodel). Formally, each semantic attribute acts as a labeling function (similar to $LFB$ of \model), i.e. $s_{i}: X_{j} \rightarrow \{0,1\}$. Similar to \cite{lampert2009learning}, the $LFB$ produces the final class label $y$ of the generated image $X_{j}$ by ranking the similarity scores between ground truth semantic attribute vectors of seen and zero-shot classes $\textbf{s}^{gt}$, and the semantic attribute vector $\textbf{s}^{j}$ for $X_{j}$: 
\begin{equation}
y = \arg \min_{y\in \{y_{1}, \cdots, y_{K}\}} \Big{(}(\Phi\cdot\Theta)\circ \textbf{s}^{j})\cdot(\textbf{s}^{gt}_{y})\Big{)}
\label{eq:zero_shot_LFB}
\end{equation}
\noindent where $(\Theta,\Phi)\sim G$ is sampled from the generator $G$ of \model, and $\circ$ denotes the Hadamard product. Following \cite{lampert2009learning}, we assume access to a deterministic semantic attribute vector, $\textbf{s}^{gt}$ (ground truth), for each seen and zero-shot class.

The adversarial framework of \zeromodel learns a non-linear mapping between text and labeled image space:
\begin{equation}
\begin{split}
& \min \max V(G,D_{image},D_{label}) = \mathbb{E}_{(X^{seen},t^{seen}) \sim \mathcal{D}_{G}} \log D_{image}(X^{seen} | \psi(t^{seen}))\\
& + \mathbb{E}_{\textbf{z} \sim \mathcal{N}(0,I)} \log(1- D_{image}(G(\textbf{z},\psi(t^{seen})))) \\
& + \mathbb{E}_{(y^{seen},t^{seen}) \sim \mathcal{D}_{G}} \log (D_{label}(y^{seen}|\psi(t^{seen}))\\
& + \mathbb{E}_{\textbf{z} \sim \mathcal{N}(0,I)} \log(1 - D_{label}(LFB(G(\textbf{z},\psi(t^{seen}))))
\end{split}
\label{eq:zeroshot7}
\end{equation}
Since raw text can be vague and contain redundant information, we encode the raw text using a text encoder and obtain a text encoding $\psi(t)$. Our encoder $\psi(t)$ is influenced by the Joint Embedding of Text and Image (JETI) encoder proposed by Reed \textit{et al.} in \cite{reed2016generative}.

At inference time, the raw text from either seen or zero-shot class is first encoded by the text encoder $\psi(t), \text{where, } t\in\{zero,seen\}$. The \zeromodel gets the noise vector $\textbf{z}\sim\mathcal{N}(0,I)$ and encoded text $\psi(t)$ as input and provides image $\tilde{X}$ and dependency parameters $\{\Theta, \Phi\}$ as outputs. Following Equation \ref{eq:zero_shot_LFB}, the $LFB(.)$ provides class label $\tilde{y}$ by aggregating the decisions of semantic attributes.
\begin{algorithm}
\SetAlgorithmName{Labeling Function}{}{}
\SetAlgoLined
\textbf{Input:} Image, Number of Clusters = $k$ (equal to number of classes $y$)\\
\textbf{Output:} Probabilistic label vector\\
\tcc{\footnotesize{Unsupervised deep learning based labeling function}}
$m$ = Num of kernels from fifth layer of pre-trained AlexNet trained using DeepCluster method \cite{caron2018deep}\;
\For{i=1$\cdots$ n}{
	\For{j = 1 $\cdots$ m}{
        $v_{avg_{ij}}$ = average value of Frobenius norm of activation map of $j^{th}$ kernel on subset of training samples from k$^{th}$ cluster\;
    }
}
\For{j = 1 $\cdots$ m}{
        $v_{Image_{j}}$ = value of Frobenius norm of activation map of $j^{th}$ kernel on Image\;
    }
return OneHot$\Big(\arg\min_i \Big[|v_{avg_{ij}} - v_{Image_{j}}|\Big]\Big)$
\caption{Labeling function based on Deep Feature}
\label{alg:HLF4}
\end{algorithm}
\vspace{-0.5cm}
\paragraph{5.4.1.1. Adding Cycle Consistency Loss to Semantic Attributes:} Since zero-shot classes are not present at training time and \zeromodel is optimized only using seen classes, we often observed a strong bias to seen classes at inference time. Such biases also been reported in earlier efforts such as \cite{zhang2019co} and we imposing an additional cycle-consistency loss to the \zeromodel, i.e.:
\begin{equation}
    \begin{split}
        & V_{zero}(G,D_{Image},D_{Label})= V(G,D_{Image},D_{Label})\\
        & + \lambda\mathbb{E}_{\textbf{s}\sim LFB(G(\textbf{z},\textbf{s}^{gt}))}||\textbf{s}-\textbf{s}^{gt}||_{1}
    \end{split}
\label{eqn_zero_lambda}
\end{equation}

\noindent in addition to \zeromodel, the generator $G(.)$ gets the $\textbf{s}^{gt}$ of seen and zero-shot classes, the noise vector $\textbf{z}\sim\mathcal{N}(0,I)$ as input to learn dependency parameters $\{\Theta, \Phi\}$ (as opposed to the usual text embedding $\psi(t)$ with $\textbf{z}\sim\mathcal{N}(0,I)$ described in Equation \ref{eq:zeroshot7}). Here, $\lambda$ is a hyperparameter, which we vary and report results in Figure \ref{fig:zero_shot}(d).
\section{Experiments and Results}
\label{sec_expts}
\subsection{Dataset}
\begin{table}
\centering
\begin{tabular}{lllll}
\cline{2-5} \cline{2-5}
\multicolumn{1}{l}{} & SVHN & CIFAR & LSUN & CHEST \\ \hline \hline
\multicolumn{1}{c}{Heuristic} & \multicolumn{3}{c}{\begin{tabular}[c]{@{}c@{}}Long edges \cite{mandal2011handwritten, alfonseca2012pattern}, \\PatchMatch \cite{barnes2009patchmatch}, \\ Local Dense \\Features \cite{sahu2018reducing}, \\ Holistic spatial \\envelope \cite{oliva2001modeling}, \\Domain\\ Knowledge \\from Experts\end{tabular}} & \begin{tabular}[c]{@{}c@{}}Rib\\border \cite{plourde2006semi, vogelsang1998detection},\\ Parametric \\model \cite{toriwaki1973pattern, van2001computer},\\ Statistical\\shape \\model \cite{ougul2015unsupervised},\\ Domain \\Knowledge\\ from Experts\end{tabular} \\ \hline
\multicolumn{1}{c}{\begin{tabular}[c]{@{}c@{}}Image\\ Processing\end{tabular}} & \multicolumn{4}{c}{\begin{tabular}[c]{@{}c@{}}SIFT \cite{lowe2004distinctive}, k-means clustering \cite{lin2010power}, \\ Bags of Keypoints \cite{csurka2004visual}\end{tabular}} \\ \hline
\multicolumn{1}{c}{\begin{tabular}[c]{@{}c@{}}Deep \\ Learning\end{tabular}} & \multicolumn{4}{c}{\begin{tabular}[c]{@{}c@{}}DeepCluster \cite{caron2018deep}, Deep features \\from \cite{xie2016unsupervised,yang2016joint}, Deep Representation \\from \cite{bautista2016cliquecnn, dosovitskiy2014discriminative, liao2016learning},\end{tabular}} \\ \hline \hline
\end{tabular}
\caption{Description of Labeling Functions used for SVHN \cite{SVHN}, CIFAR 10 \cite{CIFAR10}, LSUN \cite{LSUN} and CHEST-Xray-14 \cite{CHEST} dataset.}
\label{table_labelingfunctions}
\end{table}
\noindent\textit{6.1.1. \model and \ssmodel}: We validated \model and \ssmodel on the following datasets: (i) SVHN \cite{SVHN}; (ii) CIFAR-10 \cite{CIFAR10}; (iii) LSUN \cite{LSUN}; and (iv) CHEST, i.e. Chest-Xray-14 \cite{CHEST} datasets. For cross-domain multi-task learning and transfer learning using \model, we validated on: (i) digit dataset MNIST \cite{MNIST} and SVHN \cite{SVHN}; (ii) cloth dataset: Fashion MNIST (FMNIST) \cite{Fashion-MNIST} and LookBook \cite{yoo2016pixel}. We grouped LookBook dataset 17 classes into 4 classes: \textit{coat, pullover, t-shirt, dress}, to match number of classes of FMNIST dataset.\\
\textit{6.1.2. \zeromodel}: We evaluated \zeromodel on: (i) CUB 200 \cite{WahCUB_200_2011}; (ii) Flower-102 \cite{Nilsback08}; and (iii) Chest-Xray-14 \cite{CHEST}. We consider Nodule and Effusion (randomly selected) as zero-shot classes of Chest-Xray-14 dataset in our experiments.
\subsection{Labeling Functions}
\label{subsec:labeling_function}
\noindent\textit{6.2.1. \model and \ssmodel} shown in Table \ref{table_labelingfunctions}: We encode distant supervision signals as a set of (weak) definitions using which unlabeled data points can be labeled. We categorized labeling functions in Table \ref{table_labelingfunctions} into three categories: (i) \textbf{Heuristic}: We collect the labeling functions based on domain heuristics such as knowledge bases, domain heuristics, ontologies. Additionally, these definitions can be harvested from educated guesses, rule-of-thumb from experts obtained using crowdsourcing. The experts were given a batch ($\simeq 4000$ images of a dataset) and asked to provide a set of labeling functions. (ii) \textbf{Image Processing}: Domain heuristics from Image processing and Computer Vision. (iii) \textbf{Deep Learning}: We collect activation maps from pre-trained deep models (deep models trained in an unsupervised manner). We show an example of labeling functions used for the SVHN dataset in Labeling Functions \ref{alg:HLF4}.\\
\textit{6.2.2. \zeromodel}: The CUB 200 dataset \cite{WahCUB_200_2011} provides a set of semantic attributes to identify a class. While, for Flower-102 and Chest-XRay-14 dataset we follow \cite{al2017automatic} to identify semantic attributes, followed by \cite{reed2016learning} and get a color-based semantic attributes.
\vspace{-0.3cm}
\subsection{Implementation Details}
\noindent\textit{\model, \ssmodel and \zeromodel}: We adopt BigGAN $128\times128$ architecture \cite{brock2018large} to implement generators and discriminators of \model, \ssmodel and \zeromodel. We slightly change the last layers of BigGAN model to produce images of intended size of a dataset. In particular, $G_{Common} + G_{Image}$ is the BigGAN generator, and $G_{Parameter}$ branch (3 Fully Connected FC layers) is forked after the ``Non-Local Block'' of the BigGAN generator. Similarly, $D_{Image}+D_{Common}$ follows BigGAN discriminator, while $D_{Label}$ branch is added after ``Non-Local Block'' of the BigGAN discriminator. We follow the official hyperparametres of BigGAN, i.e. $z=120d$, train generator for 250k iterations and 5 discriminator iterations before every generator iteration, optimizer=$\{Adam\}$, learning rate for generator is $5\cdot10^{-5}$ and $2\cdot10^{-4}$ for discriminator.
\begin{figure}[!htb]
\centering
\includegraphics[width=\textwidth, height=0.6\textwidth]{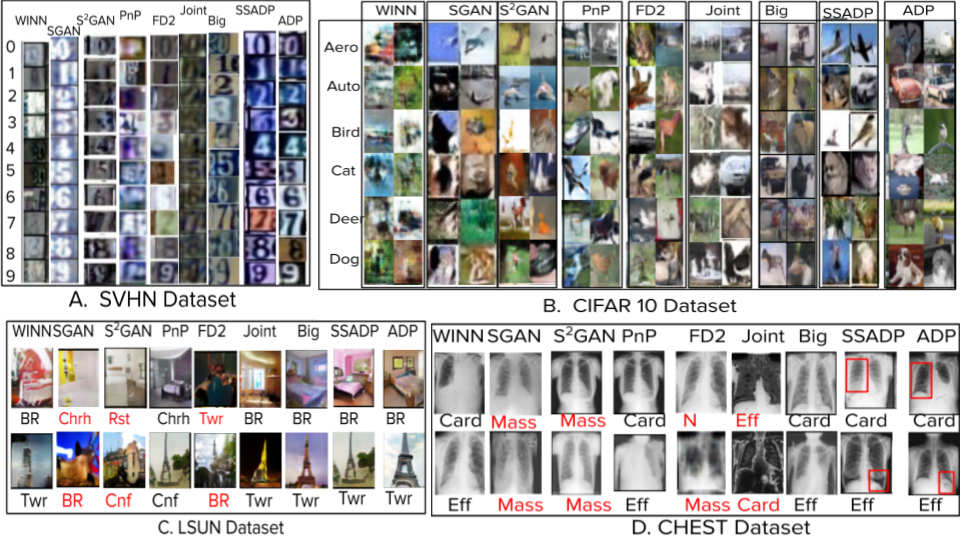}
\caption{(Best viewed in color and while zoomed) Qualitative results of different generative (both GAN and Non-GAN) methods, such as: WINN \cite{lee2018wasserstein}, Self Supervised SGAN \cite{lucic2019high}, semi-supervised S$^{2}$GAN \cite{lucic2019high}, Plug and Play PnP \cite{nguyen2017plug}, f-VAEGAN-D2 FD2 \cite{xian2019f}, JointGAN Joint \cite{pu2018jointgan}, BigGAN Big \cite{brock2018large} and our proposed \ssmodel and \model, are given in columns. (\textbf{A}) \textbf{SVHN}-(\textbf{B}) \textbf{CIFAR 10}: First columns represent class labels. We use abbreviation: Automobile ``Auto'', Aerospace ``Aero''; (\textbf{C}) \textbf{LSUN}: We use abbreviations: Tower ``Twr'', Church ``Chrh'', Bridge ``BR'', Conference Room ``Cnf'', Restaurant ``Rst''. (\textbf{D}) \textbf{CHEST}: We use abbreviation, such as: Effusion ``Eff'', Nodule ``N'', Cardiomegaly ``Card'' etc. We show wrong labels in color red and correct labels as black. While, labels are provided as input in some methods (WINN, PnP, BigGAN), our proposed methods \model and \ssmodel generates labeled images. On CHEST generated images, we additionally get disease location mark (shown as red boxes) from experts on generated images.}
\label{fig:Qualitative_comparison}
\end{figure}
\subsection{Qualitative Results Comparison with Prior Methods} 
\label{subsec:qualitative_result}
\noindent\textit{6.4.1. \model and \ssmodel}: We compared our proposed \ssmodel and \model methods against other generative (both GAN and Non-GAN) methods, such as: WINN \cite{lee2018wasserstein}, Self Supervised SGAN \cite{lucic2019high}, semi-supervised S$^{2}$GAN \cite{lucic2019high}, Plug and Play PnP \cite{nguyen2017plug}, f-VAEGAN-D2 FD2 \cite{xian2019f}, JointGAN Joint \cite{pu2018jointgan}, BigGAN Big \cite{brock2018large}, and show results in Figure \ref{fig:Qualitative_comparison}. We discuss our improved results in terms of: (i) \textbf{Image Quality}: The results of ADP and SS-ADP show a significant improvement with respect to the baseline BigGAN method. We observe, the image ``style'' (i.e. background, illumination etc.) and ``content'' (i.e. object shape, orientation) are capture thoroughly in \model and \ssmodel. The improvements in \model and \ssmodel are likely due in part to the fact that $LFB(.)$ acts as a regularizer in the training objective of generator in Algorithm \ref{alg:main} encourages \model and \ssmodel to capture modes, thus resulting an improved ``style''ness and ``content''ness. For example: in CIFAR 10, we can observe clear automobile structure and color variation for CIFAR 10 automobile ``Auto'' generated images (see Figure \ref{fig:Qualitative_comparison}).

\begin{figure}[!htb]
\centering
\includegraphics[width=\textwidth, height=0.5\textwidth]{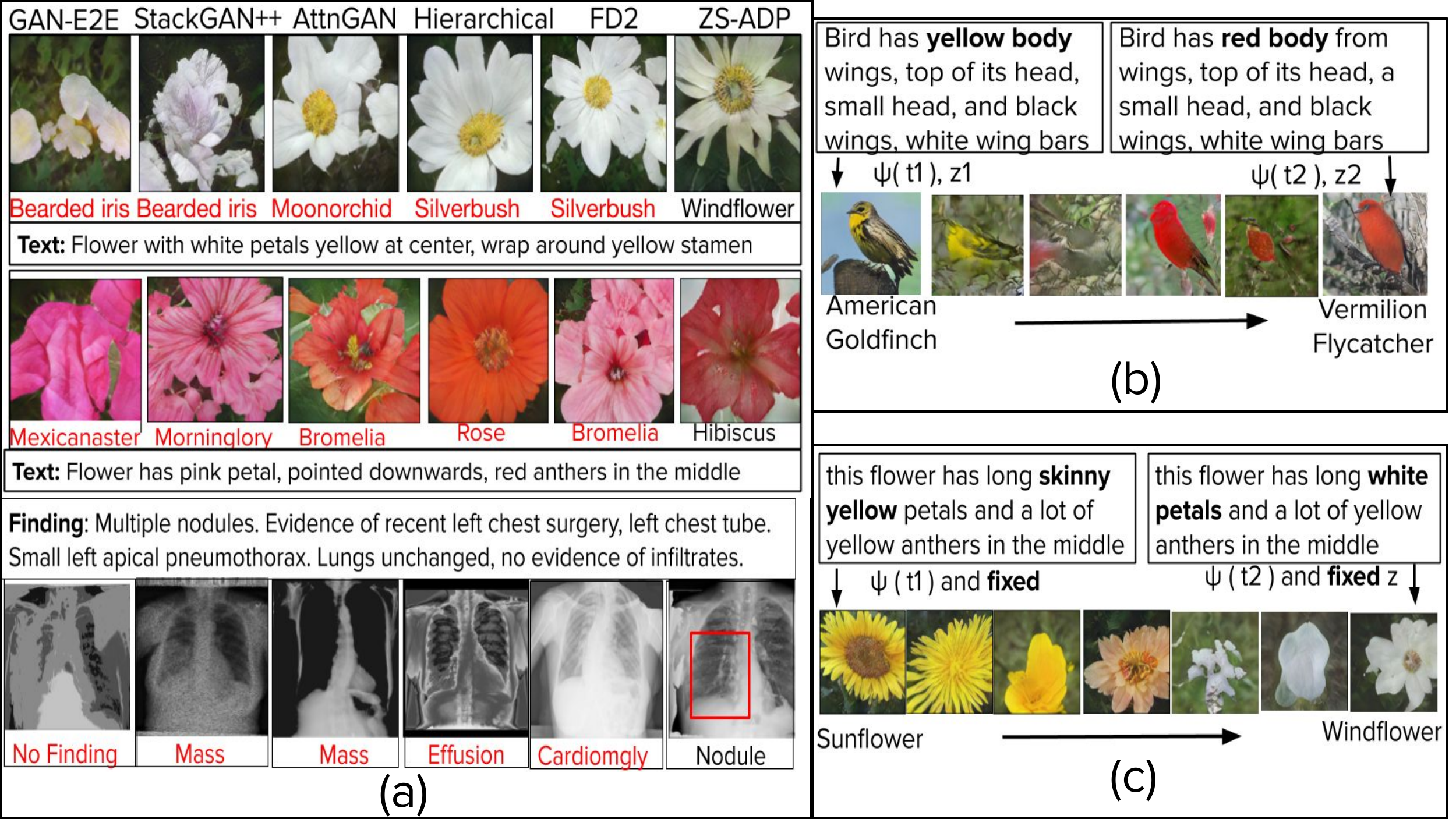}
\vspace{-0.3cm}
\caption{(Best Viwed in color, and, please zoom to see details) Generalized zero-shot labeled image generation on Flower-102, CUB 200 and Chest-Xray-14 datasets. \textbf{(a)} Generated labeled images by GAN-E2E \cite{reed2016generative}, StackGAN++ \cite{zhang2017stackgan++}, AttnGAN \cite{xu2018attngan}, Hierarchical \cite{zhang2018photographic}, FD2 (that is f-VAEGAN-D2) \cite{xian2019f} and our method \zeromodel. Existing state-of-the-art methods fail to capture image-to-label correspondence of zero-shot classes, and converge only to seen class labels (red in color). Only \zeromodel gives quality image and good image-label correspondence. \textbf{(b)} Changing text and noise vector: Generated images not only have different background but the object color also changed, and hence a change in label. \textbf{(c)} Change in text but fixed noise vector: Style of one class (i.e. Sunflower) is transferred to another class (i.e. Windflower) by changing content information.}
\label{fig:zero_shot}
\vspace{-0.4cm}
\end{figure}
We note a good variation in background, orientation, and low-level object structure on generated images by \model and \ssmodel. Similarly, on CHEST dataset, we observe low-level details, such as: exact lungs location, disease mark (shown in red box) etc. on generated images, shown in Figure \ref{table:zero_shot_quantity_result}. However, all other methods fail to capture such disease marks and generate the global image of chest X-ray. (ii) \textbf{Image to Label Correspondence}: We observe a good image-to-label correspondence (see Figure \ref{fig:Qualitative_comparison}), thus proving our claim of using labeling functions within adversarial framework of GAN architecture.

\begin{table}[]
\centering
\begin{adjustbox}{width=\textwidth}
\begin{tabular}{lllllllllllllllll}
\hline\hline
\multirow{2}{*}{Methods} & \multicolumn{4}{c}{SVHN} & \multicolumn{4}{c}{CIFAR10} & \multicolumn{4}{c}{LSUN} & \multicolumn{4}{c}{CHEST}\\
\cline{2-17} 
 &MIS&FID&$C_{RT}$&$C_{RG}$&MIS&FID&$C_{RT}$&$C_{RG}$&MIS&FID&$C_{RT}$&$C_{RG}$&MIS&FID&$C_{RT}$&$C_{RG}$\\
 & ($\uparrow$)&($\downarrow$)& ($\uparrow$)& ($\uparrow$)& ($\uparrow$)&($\downarrow$)&($\uparrow$)&($\uparrow$)& ($\uparrow$)&($\downarrow$)& ($\uparrow$)&($\uparrow$)&($\uparrow$)&($\downarrow$)&($\uparrow$)&($\uparrow$)\\ \hline\hline
WINN & 1.21 & 21.72 & 56 & 43 & 0.73 & 28.92 & 54 & 30 & 1.01 & 19.05 & 62 & 42 & 1.08 & 18.92 & 58 & 46 \\ \hline
SGAN & 1.33 & 18.03 & 51 & 48 & 1.94 & 15.91 & 38 & 30 & 1.13 & 18.72 & 68 & 60 & 2.01 & 11.84 & 71 & 64 \\ \hline
SGAN & 2.03 & 10.07 & 62 & 58 & 1.37 & 17.93 & 54 & 43 & 3.18 & 7.71 & 75 & 64 & 3.49 & 6.29 & 72 & 71 \\ \hline
PnP & 1.12 & 17.94 & 53 & 48 & 0.92 & 27.60 & 57 & 49 & 2.32 & 10.41 & 53 & 52 & 3.06 & 7.64 & 73 & 60 \\ \hline
FD2 & 1.83 & 16.73 & 72 & 70 & 1.63 & 17.02 & 58 & 50 & 2.81 & 10.90 & 73 & 62 & 2.89 & 13.63 & 69 & 58 \\ \hline
Joint & 1.84 & 13.71 & 74 & 71 & 1.17 & 18.81 & 66 & 61 & 1.91 & 14.19 & 61 & 51 & 2.31 & 9.90 & 74 & 73 \\\hline
Big & 3.04 & 8.01&75&67&2.44&13.47&67&49&3.97&7.72&73&63&3.31&6.17&74&63\\ \hline
\textbf{SSADP} & \textit{3.51} & \textit{8.32} & \textit{79} & \textit{74} & \textit{1.61} & \textit{12.91} & \textit{69} & \textit{67} & \textit{4.02} & \textit{6.41} & \textit{74} & \textit{71} & \textit{3.62} & \textit{6.01} & \textit{72} & \textit{71} \\ \hline
\textbf{\model} & \textbf{3.74} & \textbf{7.29} & \textbf{83} & \textbf{81} & \textbf{2.82} & \textbf{9.21} & \textbf{72} & \textbf{70} & \textbf{4.81} & \textbf{5.37} & \textbf{88} & \textbf{87} & \textbf{4.01} & \textbf{5.25} & \textbf{82} & \textbf{80} \\ \hline\hline
\end{tabular}
\end{adjustbox}
\caption{Qualitative results of generative (both GAN and Non-GAN) methods: WINN \cite{lee2018wasserstein}, Self Supervised SGAN \cite{lucic2019high}, Semi-supervised S$^{2}$GAN \cite{lucic2019high}, Plug and Play PnP \cite{nguyen2017plug}, f-VAEGAN-D2 FD2 \cite{xian2019f}, JointGAN Joint \cite{pu2018jointgan}, BigGAN Big \cite{brock2018large}, \ssmodel and \model, are given in columns: (i) \textbf{MIS} ($\uparrow$) Modified Inception Score \cite{santurkar2018classification}: (Higher value is better) Though our basic setup is based on Big framework, we observe almost 1 unit performance boost for \model. Similarly, the \ssmodel outperforms Big method all cases showing the efficacy of our proposed method on labeled image generation; (ii) \textbf{FID} ($\downarrow$) Frechet Inception Distance \cite{heusel2017gans}: (Lower value is better) Lower values of \ssmodel and \model w.r.t other methods on FID imply that the generated labeled images are both good in quality and produce versatile labeled image samples; (iii) \textbf{C$_{RT}$} ($\uparrow$): (Higher value is better) Top-1 classification accuracy percentage (values are in \%) of a ResNet-50 classifier trained on real labeled images and tested on generated images; (iV) \textbf{C$_{RG}$} ($\uparrow$): (Higher value is better) Top-1 classification accuracy percentage (values are in \%) of a ResNet-50 classifier trained on generated labeled images and tested on real images.}
\label{table:qualitative_results}
\end{table}

\begin{table}
\centering
\begin{tabular}{llll}
\hline\hline
\multicolumn{4}{c}{A. Image to Label Correspondence (HTT/C$_{RT}$)} \\ \hline
Methods & \multicolumn{1}{c}{\begin{tabular}[c]{@{}c@{}}Flower 102\end{tabular}} & \multicolumn{1}{c}{\begin{tabular}[c]{@{}c@{}}CUB 200\end{tabular}} & \multicolumn{1}{c}{\begin{tabular}[c]{@{}c@{}}Chest Xray\end{tabular}} \\ \hline \hline
GAN E2E \cite{reed2016generative}& (5.38/53) & (5.91/58) & (4.21/59)  \\ \hline
StackGAN++ \cite{zhang2017stackgan++}& (6.21/61) & (5.84/64) & (5.18/62)  \\ \hline
AttnGAN \cite{xu2018attngan}& (7.91/65) & (7.42/62) & (7.27/67)  \\ \hline
Hierarchical \cite{zhang2018photographic}&(8.41/68) & (8.46/69) & (8.01/68)  \\ \hline
FD2 \cite{xian2019f} & (8.72/70) & (8.79/72) & (8.67/72)  \\ \hline
\zeromodel & (\textbf{9.28/78}) & (\textbf{9.11/76}) & (\textbf{9.16/79})  \\ \hline \hline
\multicolumn{4}{c}{B. Image Quality (FID Score)} \\ \hline
Methods & \multicolumn{1}{c}{\begin{tabular}[c]{@{}c@{}}Flower 102\end{tabular}} & \multicolumn{1}{c}{\begin{tabular}[c]{@{}c@{}}CUB 200\end{tabular}} & \multicolumn{1}{c}{\begin{tabular}[c]{@{}c@{}}Chest Xray\end{tabular}} \\ \hline \hline
GAN E2E \cite{reed2016generative}& 12.88 & 12.71 & 14.82  \\ \hline
StackGAN++ \cite{zhang2017stackgan++}& 7.67 & 7.23 & 6.18  \\ \hline
AttnGAN \cite{xu2018attngan}& 5.91 & 5.42 & 5.27  \\ \hline
Hierarchical \cite{zhang2018photographic}&4.41 & 4.46 & 4.91  \\ \hline
FD2 \cite{xian2019f} & 4.32 & 4.02 & 3.67  \\ \hline
\zeromodel & \textbf{3.28} & \textbf{3.11} & \textbf{3.16}  \\ \hline \hline
\end{tabular}
\caption{\textbf{(A) Image to label correspondence} of zero-shot classes by \zeromodel. We report results in the form HTT/C$_{RT}$ to show Human Turing Test (HTT) of labeled image, and, C$_{RT}$ is the Top-1 classification score of ResNet50 classifier trained on real dataset and tested on the generated dataset; \textbf{(B) Image Quality:} The quality of images are evaluated using the FID score \cite{heusel2017gans}.}
\label{table:zero_shot_quantity_result}
\end{table}

\noindent\textit{6.4.2 \zeromodel}: We compared our proposed \zeromodel against five state-of-the-art methods: Reed \textit{et al.} (GAN-E2E) \cite{reed2016generative}, StackGAN++ \cite{zhang2017stackgan++}, AttnGAN (that is AttentionGAN) \cite{xu2018attngan}, Hierarchical (that is Hierarchical text-to-image synthesis) \cite{zhang2018photographic}, and FD2 (that is f-VAEGAN-D2) \cite{xian2019f}. Due to the unavailability of \textit{generalized zero-shot text to labeled image generation} methods, we modified the generators of \cite{reed2016generative, zhang2017stackgan++, xu2018attngan, zhang2018photographic, xian2019f} in a way that the last layers of those generators now generate both the image $X$ and class label $y$.

\begin{figure}
\centering
\includegraphics[width=0.47\textwidth, height=0.5\textwidth]{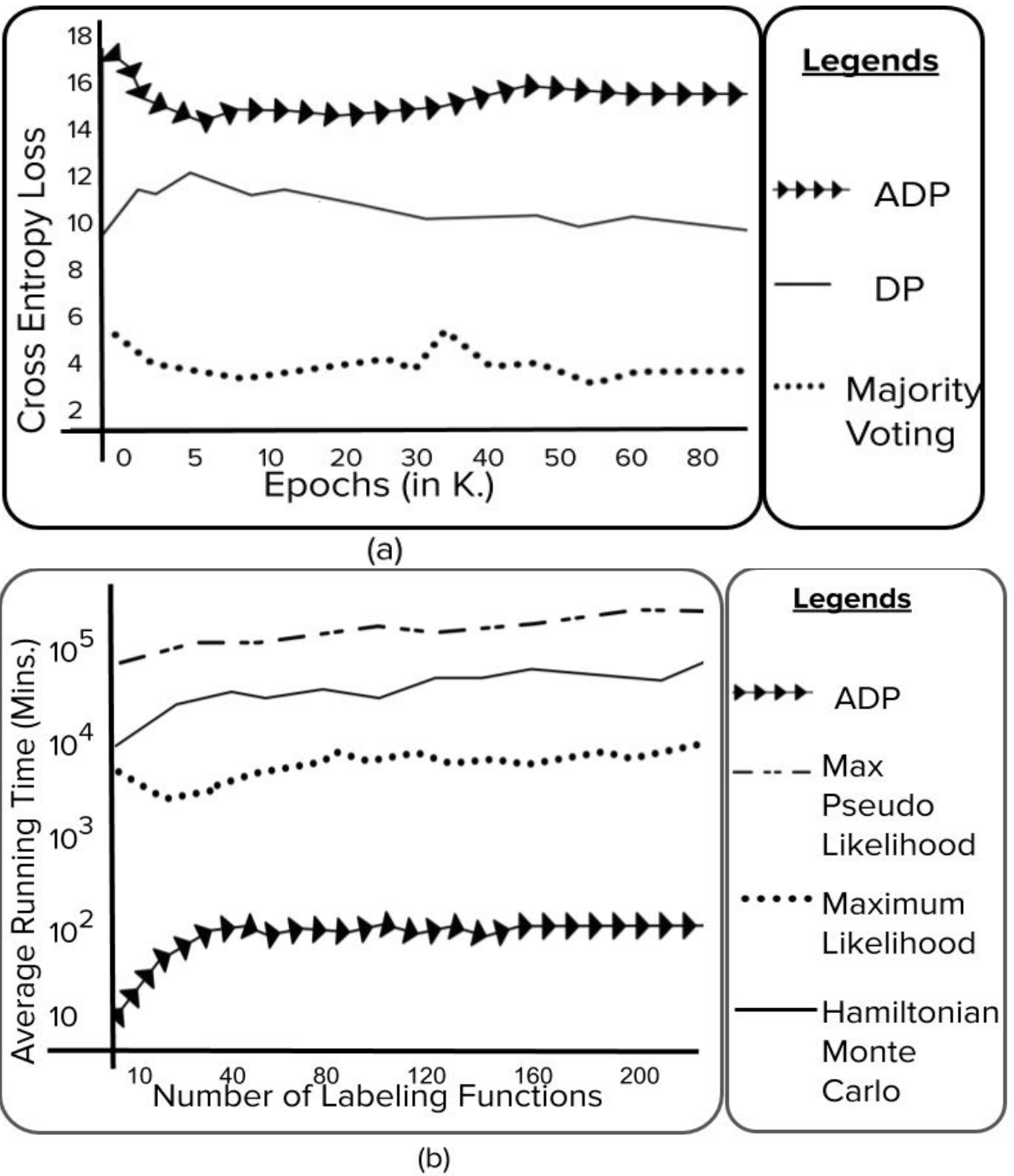}
\caption{(a) Test-time classification cross-entropy loss of a pre-trained ResNet model on image-label pairs generated by \model, \model (only its Image-GAN component) with majority voting, and \model (only its Image-GAN component) with DP for labels; (b) Average running time (in Mins.) of \model against other methods to estimate relative accuracies and inter-function dependencies.}
\label{fig:Cross}
\end{figure}

In our experiment, we fixed the $\lambda=0.3$ of Equation \ref{eqn_zero_lambda} and get the generated labeled images from text from \zeromodel. The \zeromodel generates good quality images as well as good image-label correspondence, see Figure \ref{fig:zero_shot}(a). We show latent space interpolation in Figure \ref{fig:zero_shot}(b)-(c). Figure \ref{fig:zero_shot}(b) show labeled images where we changed the noise distribution $\textbf{z}\sim\mathcal{N}(0,I)$ and the textual description. While, Figure \ref{fig:zero_shot}(c), we fixed the noise distribution $\textbf{z}\sim\mathcal{N}(0,I)$ but change the text, and we see minimal change in style, i.e. background or orientation, but a change in color and shape.
\subsection{Quantitative Result Comparison with prior methods}
\label{subsec:quantitative_result}

\begin{table}
\begin{center}
\begin{tabular}{lllll}
\hline\hline
\# LFs. & SVHN & CIFAR10 & Chest & LSUN\\
\hline\hline
3 & 10.23\% & 21.02\% & 27.39\% & 23.82\% \\ \hline
\textit{5} & 8.32\% & 8.53\% & 21.31\% & 18.30\% \\ \hline
10 & \textbf{1.40}\% & \textbf{4.81}\% & \textbf{17.93}\% & \textbf{11.62}\% \\ \hline
15 & 1.33\% & 4.92\% & 18.93\% & 13.05\% \\ \hline
20 & 1.34\% & 4.80\% & 18.45\% & 12.83\% \\ \hline
25 & 1.31\% & 4.73\% & 18.43\% & 12.82\% \\ \hline\hline
\end{tabular}
\end{center}
\caption{Performance of \model when number of labeling functions is varied.}
\label{table:optimum_number_of_labeling_functions}
\end{table}

\noindent\textit{6.5.1. \model and \ssmodel}: For the sake of quantitative comparison among \model, \ssmodel and other generative methods, we adopted four evaluation metrics (studied in \cite{lucic2019high, pu2018jointgan}): (i) \textbf{MIS} ($\uparrow$) Modified Inception Score proposed in \cite{santurkar2018classification} computes $\exp\{\mathbb{E}_x[\textbf{KL}(p(y|X)||p(y))]\}$, where $X$: image, $p(y|X)$: softmax output of a ResNet-50 classifier (trained on real labeled images), and $p(y)$: label distribution of generated samples; (ii) \textbf{FID} ($\downarrow$) Frechet Inception Distance (FID) proposed in  \cite{heusel2017gans} computes $FID(X,X_{g})=||\mu_{X}-\mu_{X_{g}}||^{2}_{2} + \text{Tr}(\Sigma_{X} + \Sigma_{X_{g}} - 2(\Sigma_{X}\Sigma_{X_{g}})^{\frac{1}{2}})$, where, $\mu$, $\Sigma$ are mean and co-variance, and $X$: real image $X_{g}$: generated image; (iii) \textbf{C}$_{RT}(\uparrow)$, i.e. Top-1 classification accuracy (in $\%$) of a ResNet-50 classifier trained on real labeled images and tested on generated images; and (iv) \textbf{C}$_{RG}(\uparrow)$, i.e. Top-1 classification accuracy (in $\%$) of a ResNet-50 classifier trained on generated labeled images and tested on real images. As a baseline the BigGAN labeled image generator obtains MIS sores: 3.04 on SVHN, 2.44 on CIFAR 10, 3.97 on LSUN, 3.31 on CHEST datasets. While, FID scores: 8.01 on SVHN, 13.47 on CIFAR 10, 7.72 on LSUN and 6.17 on CHEST datasets. 

In \model, we observe that the generated images able to achieve an average of $\sim 0.7$ units of MIS and $\sim 2$ units of FID boosts w.r.t the baseline BigGAN method. On the other hand, we observe a smaller gap between \textbf{C}$_{RT}$ and \textbf{C}$_{RG}$, suggesting that the ResNet-50 classifier of \textbf{C}$_{RT}$ can classify well the generated images, and the ResNet-50 classifier of \textbf{C}$_{RG}$ trained on generated labeled image of \model can classify real images (the generated images are good in quality and have high image-to-label correspondence). We note that, though the BigGAN secured a good \textbf{C}$_{RT}$ score but fails to perform well at \textbf{C}$_{RG}$ score across all datasets. Such observation directly implying the advantage of using $LFB(.)$ to infer labels of generated images, within the adversarial framework of baseline BigGAN.

While in \ssmodel, we note that the alternative approach of getting ``distant supervised'' labels of unlabeled image $X$ (as discussed in Section \ref{sec_extendability}), by using labeling functions of $LFB^{1}(X)$, is complementary to the baseline BigGAN model which is trained on real labeled images. The ``SSADP'' row on Table \ref{table:qualitative_results} shows the experimental results. In particular, an improvement of MIS, FID, \textbf{C}$_{RT}$ and \textbf{C}$_{RG}$ w.r.t imply show the efficacy of the \ssmodel method.

\begin{figure}
\centering
\includegraphics[width=0.57\textwidth,height=0.2\textwidth]{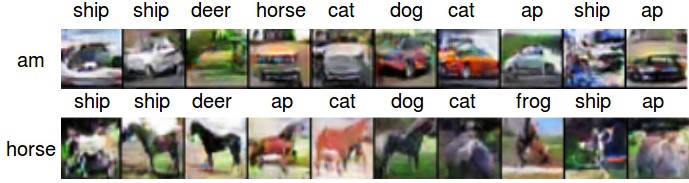}
\caption{Sample results of image-label pairs generated by combining a vanilla GAN (for image generation) and DP \cite{ratner2016data} (for label generation) using the same labeling functions used in this work. Row labels represent the original class label (am = automobile) and column labels are provided by DP. Note the poor image-label correspondence, supporting the need for our work.}
\label{fig:DP_Label}
\end{figure}

6.5.2.\textit{ZS-ADP}: ZS-ADP is evaluated using three evaluation metrics, such as: (i) Image-Label Correspondence of Zero-Shot Classes; and (ii) Image quality, and the results are shown in Table \ref{table:zero_shot_quantity_result} (A) and (B).\\
6.5.2.1.\textit{Image to label Correspondence of Zero Shot Classes}: We performed: (i) HTT, i.e. Human Turing Test: 40 experts were asked to rate the image to label correspondence of 400 zero shot labeled images. Experts were given a score on a scale of 1-10, and the aggregated result is shown in Table \ref{table:qualitative_results}. (ii) $C_{RT}$, i.e. Classification Score: Top-1 classification performance of ResNet-50 classifier trained on real labeled images and tested on generated zero-shot labeled images.

2. \textit{Image Quality}: Image quality is evaluated using the FID score \cite{heusel2017gans}. The results are shown in Table \ref{table:zero_shot_quantity_result} (B). We observed our method performs fairly well and secures a good FID for image quality.
\section{Discussion and Analysis}
\label{sec_analysis}
\subsection{Optimal Number of Labeling Functions}
\label{subsec_num_of_labeling_functions_analysis}
We trained \model using different number of labeling functions. Table \ref{table:optimum_number_of_labeling_functions} suggests that 10-15 labeling functions provides the best performance. We report the test time cross-entropy error of a pretrained ResNet model with image-label pairs generated by \model.
\vspace{-0.3cm}
\subsection{Comparison against Vote Aggregation Methods}
We compared \model, both with majority voting and Data Programming (DP, \cite{ratner2016data}). We studied the test-time classification cross-entropy loss of a pre-trained ResNet model on image-label pairs generated by \model, \model (only its Image-GAN component) with majority voting and DP. The results are presented in Figure \ref{fig:Cross}a.
\vspace{-0.3cm}
\subsection{Adversarial Data Programming vs MLE-based Data Programming:}
To further quantify the benefits of our \model, we also show how our method compares against Data Programming (DP) \cite{ratner2016data} using different variants of MLE: MLE, Maximum Pseudo likelihood, and Hamiltonian Monte Carlo. Figure \ref{fig:Cross}b presents the results and shows that \model is almost 100X faster than MLE-based estimation. Figure \ref{fig:DP_Label} also shows sample images generated by the vanilla GAN, along with the corresponding label assigned by MLE-based DP using the same labeling functions as used in our work.
\section{Conclusions}
Paucity of large curated hand-labeled training data forms a major bottleneck in deploying machine learning methods in practice on varied application domains. Standard data augmentation techniques are often limited in their scope. Our proposed Adversarial Data Programming (\model) framework learns the joint data-label distribution effectively using a set of weakly defined labeling functions. The method shows promise on standard datasets, as well as in transfer learning and multi-task learning. We also extended the methodology to a generalized zero-shot labeled image generation task, and show its promise. Our future work will involve understanding the theoretical implications of this new framework from a game-theoretic perspective, as well as explore the performance of the method on more complex datasets.
\bibliography{elsarticle-template}
\end{document}